\documentclass[conference]{IEEEtran}
\IEEEoverridecommandlockouts

\usepackage{cite}
\usepackage{amsmath,amssymb,amsfonts}
\usepackage{algorithmic}
\usepackage{graphicx}
\usepackage{subcaption}
\usepackage{textcomp}
\usepackage{xcolor}

\usepackage{caption}
\usepackage{subcaption}
\begin{document}

\title{SlimNets: An Exploration of Deep Model Compression and Acceleration}
\author{\IEEEauthorblockN{Ini Oguntola*, Subby Olubeko*, Christopher Sweeney*\thanks{* All three authors contributed equally to this paper.}}
\textit{Massachusetts Institute of Technology}\\
Cambridge, MA, USA \\
\{ini, subbyo, csweeney\}@mit.edu}
\maketitle

\begin{abstract}
Deep neural networks have achieved increasingly accurate results on a wide variety of complex tasks. However, much of this improvement is due to the growing use and availability of computational resources (e.g use of GPUs, more layers, more parameters, etc). Most state-of-the-art deep networks, despite performing well, over-parameterize approximate functions and take a significant amount of time to train. With increased focus on deploying deep neural networks on resource constrained devices like smart phones, there has been a push to evaluate why these models are so resource hungry and how they can be made more efficient. This work evaluates and compares three distinct methods for deep model compression and acceleration: weight pruning, low rank factorization, and knowledge distillation. Comparisons on VGG nets trained on CIFAR10 show that each of the models on their own are effective, but that the true power lies in combining them. We show that by combining pruning and knowledge distillation methods we can create a compressed network \textit{85 times smaller than the original}, all while retaining 96\% of the original model's accuracy.
\end{abstract}

\section{Introduction}
In 1998 the state of the art convolutional net was \textit{LeNet}, a model with 60 thousand parameters \cite{lenet}. By 2014 that number had increased to 138 million, with new neural networks such as VGGNet becoming increasingly large \cite{vgg}. Today we have GPUs that can perform trillions of operations per second to help train models with millions of parameters. The development of the computational means necessary to process such huge networks has facilitated many of the advances in the field, and continues to do so today. However, in situations where resources are constrained, such as deployment on mobile devices, it is important to make these models more efficient.

In this paper we explore various approaches to deep network compression and acceleration, focusing primarily on compression as a way to make deep neural networks use fewer resources. While the methods presented in this work have been individually developed by others, we combine them in the same application for the compression of a single network and analyze the effects of different combinations of compression approaches. We show that these various methods, while individually effective, can be compounded for multiplicative gains in compression rate. With VGG19 as our reference network on the CIFAR10 dataset \cite{cifar}, we are able to create a compressed network 85 times smaller while retaining 96\% of the original test accuracy.

\section{Prior Work}

Most of the existing work in the area of model compression and acceleration falls into one of four approaches: \textit{pruning}, \textit{low-rank factorization}, \textit{transferred convolutional filters}, and \textit{knowledge distillation} (KD). We describe the general approaches in more detail throughout the rest of this section.

\subsection{Pruning}
Pruning methods for deep neural networks all focus on removing redundant or unimportant weights to create models with smaller storage sizes. Deciding which weights are unimportant is the central question in this area of research. Before the recent wave of deep convolutional neural networks, researchers investigated the effect on the loss function (via a second order Taylor approximation) when setting certain weights to zero \cite{Brain}\cite{Surgeon}. These works used such methods to determine the ``saliency" of particular weights and pruned weights of low saliency or importance. As neural networks became larger and deeper, the focus in network pruning research shifted to simply pruning weights with low absolute values \cite{connections} \cite{ExploreSparsity} \cite{prune}. The magnitudes of weights in a neural network serve as a good proxy for how much effect a particular weight has on the loss function. Equation 1 displays the classic Stochastic Gradient Descent update equation for a particular weight $w_t$ at timestamp $t$.
\begin{equation}
w_{t+1} = w_t + \alpha \nabla_w L(w_t)
\end{equation}
Weights that are important and have a greater effect on the loss function, $L(\cdot)$, will have a larger gradient term, resulting in larger magnitude changes to the weights in each training step. The higher velocity of these weights gives them a higher probability of having larger absolute values.

There are also methods that prune entire convolutional filters at a time using heuristics involving weight magnitudes of the convolution kernel and the gradients for those weights as in Molchanov et al. \cite{channelprune}. However, for clarity we choose to focus on a simpler pruning scheme described in \cite{prune}. We now describe this method in more detail.

The work by Zhu et al. \cite{prune} presents a method for gradually pruning a pretrained neural network to a layer-wise target weight sparsity (lowest magnitude weights in each layer are pruned until the desired sparsity for the layer is reached). Instead of pruning a pretrained deep neural network all at once, the authors of the paper claim that gradually pruning the weights during training allows the network to dynamically recover from lost weights. They present a pruning scheduler to calculate how many weights in a given layer to prune per training step (presented in equation 2) \cite{prune}.
\begin{equation}\resizebox{.9\hsize}{!}{$
s_t = s_f + (s_i-s_f)(1-\frac{t-t_0}{n\Delta t})^3 \textrm{ for } t \in \big\{t_0, t_0+\Delta t, ... , t_0+n\Delta t\big\}$}
\end{equation}
where $s_t$ is target sparsity for the current step, $s_f$ is the final sparsity target, $s_i$ the initial sparsity, $t$ the current time step, $t_0$ the initial training step, $n$ the number of pruning steps, and $\Delta t$ the number of training steps in between each pruning step.
They evaluate their algorithm on the Inception V3 neural network, reporting impressive compression rates for mild losses in test accuracy. Our experiments evaluate this gradual pruning approach on VGG19, reporting similar trade-offs.

\subsection{Low-Rank Factorization}
The goal of low-rank factorization methods for convolutional neural networks is to speed up and compress a network by eliminating redundancy in the 4D tensors that serve as its convolutional kernels. The method we use in our experiment relies on decomposing the tensors via SVD and using the results to compute low-rank approximations of the kernels. This approach was inspired by the paper by Tai et. al \cite{tai}, in which the authors present an algorithm to compute an exact closed form solution of the low-rank tensors. The algorithm builds a low-rank constrained network with approximated kernels derived from the weights of an unconstrained pretrained model. The way we constrain the rank of the new model is by replacing each convolutional layer from the original model with two new layers, $V$ and $H$. If the original layer has $C$ inputs and $N$ outputs, then $V$ will have $C$ inputs and $K$ outputs, while $H$ will have $K$ inputs and $N$ outputs, where $K < N$. This constrains the rank of the layer to be $K$. Each convolutional layer in the network gets a particular $K$ value which is a hyperparameter that can be learned during training. 

Next, we approximate the kernels using the scheme described by Tai et. al, which works as following:
Define $\mathcal{W}$ to be the $C \times d \times d \times N$ weight tensor associated with a given convolutional layer and $\mathcal{T}: \mathbb{R}^{C \times d \times d \times N} \rightarrow \mathbb{R}^{Cd \times dN}$ to be a bijection that maps tensor $T$ to matrix $M$ according to the rule $T_{i_1,i_2,i_3,i_4} = M_{j_1,j_2}$ where
\begin{equation}
j_1 = \left(i_1 - 1\right)d + i_2,\: j_2 = \left(i_4 - 1\right)d + i_3.
\end{equation}
Let $W := \mathcal{T}[\mathcal{W}]$ and let $W = USV^T$ be the Singular Value Decomposition of $W$. We compute the weight kernels for $V$ and $H$ according to:
\begin{equation}
\hat{\mathcal{V}}_k^c\left(j\right) = U_{\left(c-1\right)d+j,\:k} \sqrt{S_{k,k}}
\end{equation}
\begin{equation}
\hat{\mathcal{H}}_n^k\left(j\right) = V_{\left(n-1\right)d+j,\:k} \sqrt{S_{k,k}}
\end{equation}
where $\hat{\mathcal{V}}$ and $\hat{\mathcal{H}}$ are the kernels for $V$ and $H$ respectively.
Finally, we train the new low-rank model starting with the approximated weights. The many added layers in the new model make it susceptible to the problem of exploding gradients, so we make use of the technique of batch normalization on all of the $H$ layers in order to combat this. The rank-constrained model was trained and validated over 300 epochs and we kept track of the accuracy, loss, and elapsed time over each epoch. Our experimental results showed significant improvements in the memory and time consumed by the low-rank constrained model as well as improvement in test accuracy.

\subsection{Knowledge Distillation}
The basic idea behind knowledge distillation is to train a smaller student model to reproduce the output of a larger teacher model. Instead of outright training the smaller model on the original data, we aim to somehow take advantage of the function already learned by the teacher model and transfer knowledge to the student.

Most neural network classifiers have a ``softmax" output layer that produces probabilities for each class. Given a logit score $z_i$, we obtain  probability $p_i$ via the following equation:
\begin{equation}
p_i = \frac{\exp(z_i/T)}{\sum_j \exp(z_j/T)}
\end{equation}
where usually we have $T=1$.
One of the earliest examples of knowledge distillation for deep networks was suggested by Caruana et al. \cite{caurana}. They passed the input through the original model and used the output of this softmax layer as synthetic labels for a compressed model. Building upon this, Hinton et al. \cite{hinton} use a softened version of the teacher output with higher values of $T$, and train the compressed model to predict both the teacher output and the true classification labels.

One technique -- \textit{FitNets} -- proposed by Romero et al. \cite{fitnets}, tries to utilize the intermediate representations of the teacher network to train student models that are both thinner and deeper than the teacher; another technique uses conditional adversarial networks to learn the teacher function \cite{cgan}. Recent work has also tackled data-free knowledge distillation in the case where the original data is not available to student models \cite{datafree}.

In this paper we primarily focus on the approach from \cite{hinton} and variations.

\subsection{Transferred Convolutional Filters}
Transferred convolutional filters can also be used as a method of model compression. Let us have input $x$, network or layer $\Phi$, and transform matrix $\mathcal{T}$. Then the network is structure preserving or ``equivariant" if the following holds:
\begin{equation}
\Phi(\mathcal{T}x) = \mathcal{T}'\Phi(x)
\end{equation}
where $\mathcal{T}$ and $\mathcal{T}'$ are note necessarily the same. The idea proposed by Cohen et al. \cite{cf} is to build a network consisting of convolutional layers structured this way in order to facilitate weight sharing and increase the expressiveness of the network without significant extra computations overhead for translations, rotations and reflections.

One thing to note is that this approach to model compression is limited to use for models that contain only convolutional layers, whereas the other three aforementioned approaches can be used on models containing both convolutional and fully-connected layers. For this reason we do not focus on transferred convolutional filters in this work.

\section{Experiments}
We perform sparse gradual pruning, low rank matrix factorization, and knowledge distillation on a pretrained VGG19 network trained on CIFAR10. We compare the evolution of training, as well as model compression rate, inference time speed ups, and overall top-1\%-accuracies for each method. The pretrained VGG19 model we use has a test accuracy of 92.4\%, and we show that it is possible to achieve very competitive accuracies with models that are orders of magnitude smaller and quicker.

\subsection{Pruning}

\begin{table}[bp]
\caption{Pruning Results}
\begin{center}
\begin{tabular}{lc}
\textbf{Method} & \textbf{Accuracy}  \\
\hline
Non-Gradual Pruning 40\% sparse & 89.69 \\
Gradual Pruning 40\% sparse & 92.22 \\
Non-Gradual Pruning 75\% sparse & 89.13\\
\textbf{Gradual Pruning 75\% sparse} & \textbf{91.57}  \\
Non-Gradual Pruning 87.5\% sparse & 86.16\\
Gradual Pruning 87.5 percent & 88.82 \\
~\\
\multicolumn{2}{l}{* where \textbf{bold} is the best tradeoff of accuracy/model compression}
\end{tabular}
\label{tab:pruning}
\end{center}
\end{table}

\begin{figure}
\centering
\subcaptionbox{Sparsity over Training Duration}[0.49\linewidth]{\includegraphics[width=0.49\linewidth]{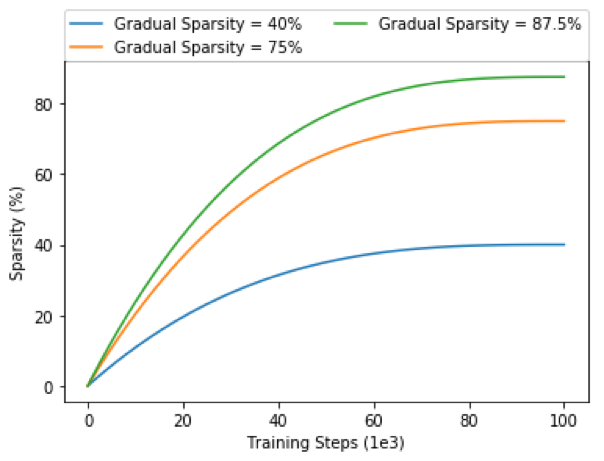}}
%\hfill
\subcaptionbox{Loss over Training Duration}[0.49\linewidth]{\includegraphics[width=0.49\linewidth]{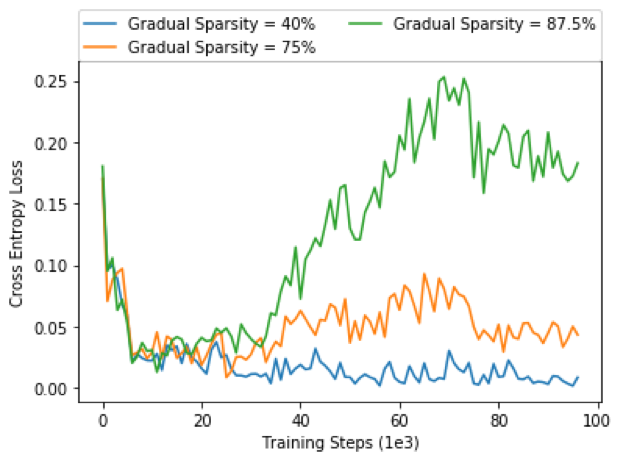}}
\caption{Evolution of Network Sparsity vs Loss for Pruning}
\label{fig:pruning_test}
\end{figure}

We prune a pretrained VGG19 network using the gradual sparse pruning algorithm presented in \cite{prune}, for 40\%, 75\%, and 87.5\% layer-wise target sparsity. We compare the resulting top-1\%-test accuracies of the gradually pruned models to the models pruned all at once then trained for the same duration (100,000 training steps). The final network accuracy results for our experiments can be found in Table \ref{tab:pruning}.

Examining the results, we clearly see that gradually introducing sparsity into the model helps to yield higher test accuracies than the non gradual pruning approach. For a 75\% gradually pruned model, we are able to achieve up to 91.57 top-1\%-test accuracy. This is about a 2.5\% increase over the same 75\% non gradually pruned model. The reasoning for the superiority of gradually pruning a network, as noted in \cite{prune}, is that gradually pruning, especially with a quicker pruning rate earlier on in training, allows the network to ``heal" from damage done before the network converges. This line of reasoning is supported by the data, but a deeper analysis into how missing weights affect the network from an optimization standpoint is needed to complete the story.

Analysis on how the loss evolves during pruning and training gives interesting insights into how parameters are used in the network. Figure \ref{fig:pruning_test} shows the evolution of sparsity and loss over time for the gradual pruning method. 

Surprisingly, when gradually pruning with a larger target sparsity, there is a point at the 60,000th training step where the loss shoots up, particularity for the 87.5\% sparse model. This characteristic was also noted in \cite{prune}. This phenomenon is most likely due to the fact that when the network settles into deeper local/global minima and relies more on certain weights, pruning these weights have larger and adverse effects on the loss function. Interestingly, the network is still able to recover a little bit for the 87.5\% sparsity and completely recover for the 75\% sparsity model. These results give an interesting insight into how the network depends on weights throughout the training process. 

Some differences were also noted between \cite{prune} and our experiments. The 87.5\% sparse model for inception V3 was able to fully recover from the sudden increase in loss during training, while our model was not very effective at recovering. Perhaps Inception V3's more nested architecture makes it more robust to parameter pruning over VGG.

\subsection{Low Rank Matrix Factorization}
\begin{figure}[tbp]
\begin{center}
\includegraphics[width=0.49\linewidth]{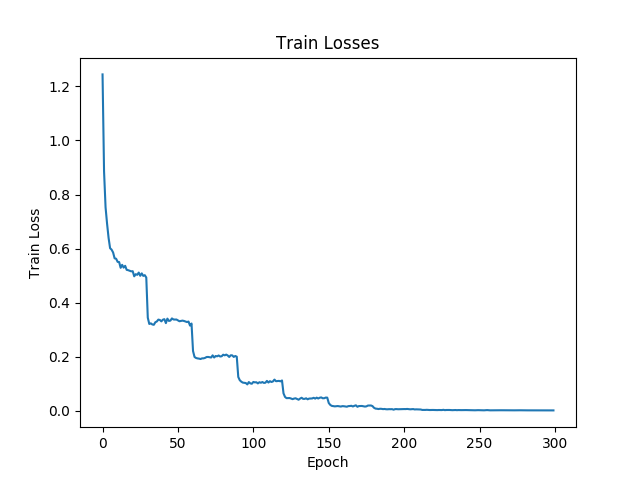}
\includegraphics[width=0.49\linewidth]{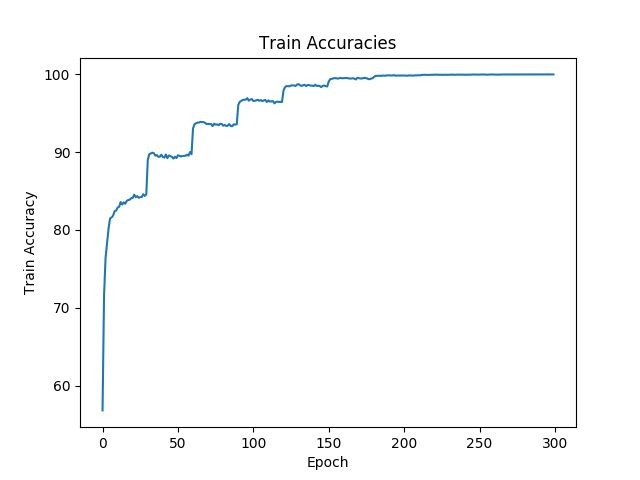}
\includegraphics[width=0.49\linewidth]{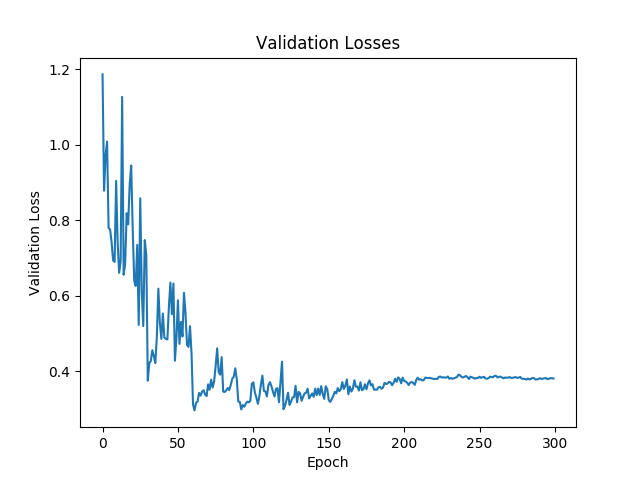}
\includegraphics[width=0.49\linewidth]{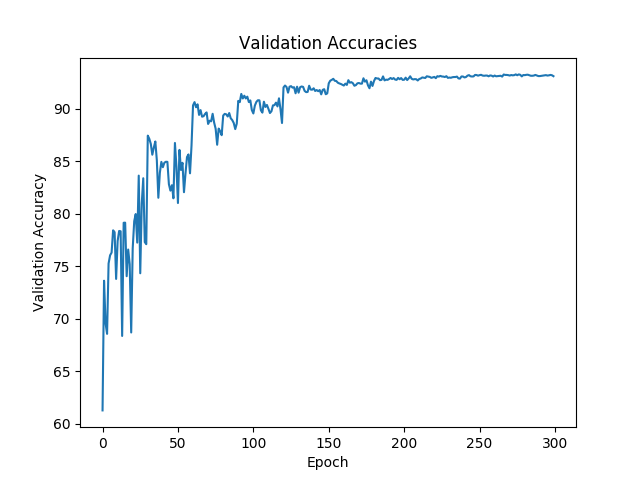}
\end{center}
\caption{Loss and Accuracy Curves for Low Rank Factorization}
\label{fig:lrf_loss_and_acc}
\end{figure}

\begin{figure}[tbp]
\begin{center}
\includegraphics[width=0.49\linewidth]{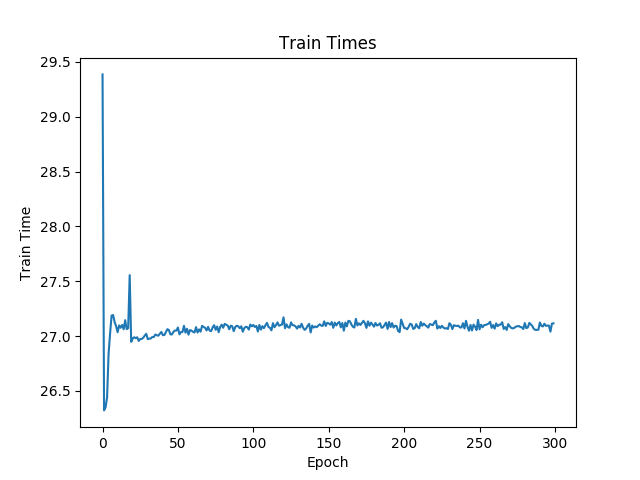}
\end{center}
\caption{Evolution of Training Time over Epochs for Low Rank Factorization}
\label{fig:lrf_train_times}
\end{figure}

We trained a low rank constrained VGG19 net with batch normalization on the CIFAR10 dataset. The loss and accuracy curves can be found in Figure \ref{fig:lrf_loss_and_acc}, and the evolution of the training time per epoch is depicted in Figure \ref{fig:lrf_train_times}. 

After 300 epochs of training, the rank-constrained VGG19 achieved a highest test accuracy of 93.44\% which surpasses the 92.43\% achieved by the unconstrained model. The inference time of the trained model over the whole dataset was 2.315 seconds compared to 2.363 seconds for the unconstrained net. Although there was little speedup in inference time, we saw that weights of the low-rank model, which used 32.3 MB on disk, were compressed to about 41.2\% of the size of those of the unconstrained model which took 78.4 MB on disk. These results align with those in the paper by Tai et. al, as the authors also saw improvement across the board in their rank-constrained model.

\subsection{Knowledge Distillation}
We take the approach detailed in \cite{hinton}, training various smaller neural networks to both approximate the output of a pretrained VGG19 model trained on the CIFAR10 dataset and to reproduce the correct labels. Input images are passed through the VGG19 model to collect the scores it assigns across all of the classes as synthetic data; this synthetic data is then used as targets for each of the smaller models.

\subsubsection{Model Architectures}
We examine six different model architectures. The first two -- ``SNN-1k" and ``SNN-10k" -- consist of two convolutional layers and a single hidden layer with 1000 and 10000 hidden units respectively. The next two -- ``CNN-1k" and ``CNN-10k" -- have an additional hidden layer with 84 hidden units inserted before the output layer. We also train on \textit{AlexNet}, a deep convolutional net with 6 convolutional layers, as detailed in \cite{alexnet}. Finally we have the deep convolutional \textit{Network in Network}, or ``NIN", consisting of 9 convolutional layers, as detailed in \cite{nin}.
\subsubsection{Training Procedure}
We train all of these models for 300 epochs, both on the original labels and with modified loss as described in \cite{hinton}.
\subsubsection{Results}

\begin{table}[tbp]
\caption{Knowledge Distillation Results}
\begin{center}
\begin{tabular}{lccc}
\textbf{Model} & \textbf{\# Parameters} & \textbf{Accuracy} & \textbf{Accuracy w/ KD} \\
\hline 
VGG19 & $\sim$ 20M & 92.4 & N/A \\
SNN-1k & $\sim$ 415k & 76.96 & 77.87 \\
SNN-10k & $\sim$ 4M & 77.17 & 79.19 \\
CNN-1k & $\sim$ 490k & 79.07 & 81.23 \\ 
CNN-10k & $\sim$ 4.9M & 80.85 & 82.23 \\
AlexNet & $\sim$ 2.5M & 72.27 & 80.78 \\
\textbf{NIN} & \textbf{$\sim$ 965k} & \textbf{88.88} & \textbf{90.57} \\
~\\
\multicolumn{4}{l}{* where \textbf{bold} is the best tradeoff of accuracy/model compression}
\end{tabular}
\label{tab:kd}
\end{center}
\end{table}
As we can see in Table \ref{tab:kd}, training with synthetically labeled data from the teacher model consistently results in improved accuracy over simply training from the original labels alone. We also see that the student network accuracy is highly dependent on model architecture; we usually see high accuracy with knowledge distillation for student models that already performed fairly well on the original data. Making nets wider by adding more units has little effect on either the original accuracy or accuracy with knowledge distillation. It is not surprising to see that, taking into consideration both accuracy and model compression, the best performing model on the original data was also best the best performing model with knowledge distillation. This best performing compressed model also resulted in a 1.15$\times$ speedup of inference time.

\section{Individual Comparison between Methods}

\begin{table}[tbp]
\caption{Comparison of Methods}
\begin{center}
\begin{tabular}{l*{4}{c}r}
\textbf{Method} & \textbf{Compression Rate} & \textbf{Speed-up} & \textbf{Accuracy} \\
\hline
Sparse Pruning & 4$\times$ & None & 91.57 \\
Low-Rank Factorization & 2.4$\times$ & 1.02$\times$ & \textbf{93.44} \\
Knowledge Distillation & \textbf{21.3$\times$} & 1.15$\times$ & 90.57 \\
\end{tabular}
\label{tab:comparison1}
\end{center}
\end{table}

In Table \ref{tab:comparison1} we stack up the model sizes and inference times for the result of each method with the best accuracy/model size trade-off. Although this trade-off can be subjective, we choose the results with the best accuracy to model size ratio.

The highest compression rate is seen in the distilled model; from a pretrained VGG19 network with a 92.4\% test accuracy we can achieve as much as a 21.3$\times$ reduction in model weights using knowledge distillation and modest speed up in runtime inference with only a few percent degradation in accuracy.

Low-rank matrix factorization was the method best at preserving the accuracy of the original model. Although the compression rate was only about 2.4$\times$, we were able to achieve a 93.44\% accuracy on the CIFAR10 test set, which is even higher than the original test accuracy.

Sparse pruning is a competitive approach, achieving 4$\times$ compression with less than a percent of test accuracy degradation. Sparse pruning, however, does not have an immediate solution for how to speed up inference time after pruning weights in the network. Unlike knowledge distillation and low-rank matrix factorization, our implementation of gradual pruning simply masks pruned weights to 0, so the architecture is not changed and thus no inference time speedup is utilized. Special hardware would be needed to take advantage of the sparsity in the pruned network. Additionally, methods for merging sparse layers into a layer that can perform the tasks of the others could speed up inference time. Although sparse pruning of networks do not offer a new model architecture with quicker inference time, this method is extremely versatile, allowing the selection of any target compression rate. We were able to achieve a compression rate higher than that of low rank factorization with an accuracy higher than that of knowledge distillation.

\section{Combining Methods}

\begin{table}[tbp]
\caption{Comparison of Method Combinations}
\begin{center}
\begin{tabular}{l*{4}{c}r}
\textbf{Methods} & \textbf{Compression Rate} & \textbf{Speed-up} & \textbf{Accuracy} \\
\hline
Pruning + KD & 21.3$\times$ & 1.15$\times$ & 89.81 \\
Low-Rank + Pruning & 9.7$\times$ & 1.02$\times$ & \textbf{91.42} \\
Low-Rank + KD & 21.3$\times$ & 1.15$\times$ & 90.08 \\
KD + Low-Rank & 50.4$\times$ & 1.05$\times$ & 89.75 \\
KD + Pruning & \textbf{85$\times$} & 1.15$\times$ & 88.69 \\
\end{tabular}
\label{tab:comparison2}
\end{center}
\end{table}

\begin{figure}
\centering
\subcaptionbox{Sparsity over Training Duration}[0.49\linewidth]{\includegraphics[width=0.49\linewidth]{sparse}}
%\hfill
\subcaptionbox{Loss over Training Duration}[0.49\linewidth]{\includegraphics[width=0.49\linewidth]{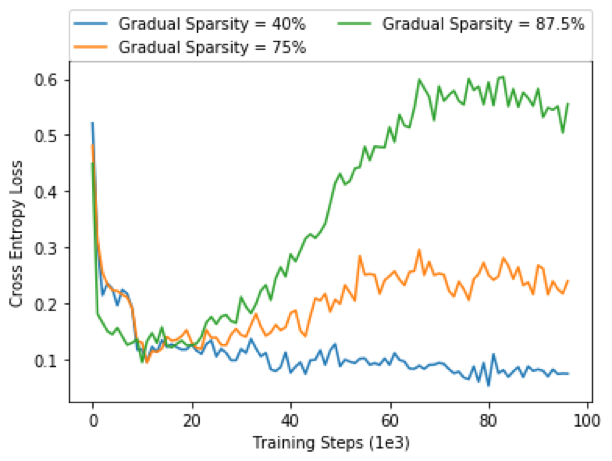}}
\caption{Evolution of Network Sparsity vs Loss for Best Knowledge Distillation Network}
\label{fig:test}
\end{figure}

We now experiment with combining methods of model compression for compound effects; the results can be found in Table \ref{tab:comparison2}.

Knowledge distillation from the pruned and low rank models was less effective than distillation from the full VGG model, which was expected. However, using pruning and low rank factorization on the already distilled model resulted in the highest rates of compression: a whopping 85$\times$ and 50.4$\times$ respectively. 

The loss curves for gradually pruning the knowledge distilled model, as shown in Figure \ref{fig:test}, are similar to those for the gradually pruned VGG19 model. There are some notable differences however. For the 87.5\% sparse model, the  network does not recover at all from the characteristic jump in loss during pruning. Additionally, the loss curves for the 40\% and 75\% curves are further apart than in the VGG19 graphs. The intuition for this observation is that the since the knowledge distilled model is more efficient in its use of weights, it has less ``pruning flexibility" or fewer weights that are meaningless to the classification  task. This explains why the 87.5\% and 75\% sparse curves do not recover like in the VGG19 case. This results in lower test accuracies after training.

The highest accuracy was obtained when we pruned the low-rank model; we were able to achieve a compression rate of 9.7$\times$ while keeping test accuracy at 91.42\% compared the original test accuracy of 92.4\%. Compare this to the 87.5\% sparse gradually pruned model in Table \ref{tab:pruning}, with a compression rate of 8$\times$ and an accuracy of 88.82\%. By combining low rank factorization with pruning we were able to obtain both a higher compression rate and a higher accuracy than with pruning alone.

\section{Discussion}

A particularly surprising result is the compound effects of combining multiple methods of model compression. By combining low rank factorization and pruning, we were able to achieve 99\% of the original test accuracy with a model 10 times smaller. By combining knowledge distillation and low rank factorization, we were able to achieve 97\% of the original test accuracy with a model 50 times smaller. Finally, by combining knowledge distillation and pruning, we were able to achieve 96\% of the original test accuracy with a model 85 times smaller.

Across the board, starting with a low-rank model resulted in the highest accuracies, while starting with a distilled model resulted in the highest compression rates; in both cases combining pruning was able to amplify these compression rates. We can take advantage of the differences in priority for the low-rank and distillation compression methods to control for the tradeoff between compression and accuracy, and the versatility of the pruning method allows us to still do this while compressing the model even further. Taken together, this facilitates very targeted model design.

More generally, our approach can be easily applied to other models and classification datasets. Pruning can be used with any model. Most classification models include a softmax output layer, which allows for knowledge distillation, and any model containing convolutional layers can benefit from low rank factorization.

\section{Conclusion}

Deep neural networks have become increasingly relevant in machine learning and in a wide array of applications. Many of these applications involve deploying deep neural networks on resource constrained devices like smart phones, but most state-of-the-art deep networks use a great deal of computational resources. Through our approach of combining model compression methods we are able to create a compressed network \textit{85 times smaller than the original}, all while retaining 96\% of the original model's accuracy.

This indicates that all of these approaches are not disparate; they can work together and be combined for some incredible rates of compression and speedup. The approaches all target different aspect of deep network inefficiency, and combined we can achieve multiplicative gains.

Finally, the results of this work indicate just how inefficient state-of-the-art deep networks can be; there is room for future work in creating better learning algorithms for smaller networks.

\section*{Acknowledgments}
We would like to thank the 6.883 staff at MIT for their instrumental feedback on this project.

\section*{Supplemental Material}
Our code for all the methods implemented in this project can be found at {https://github.com/ChristopherSweeney/SlimNets}.

\end{document}